%% file: main.tex
\documentclass{article}

     \usepackage[preprint,nonatbib]{neurips_2021}

\usepackage[utf8]{inputenc} 
\usepackage[T1]{fontenc}    
\usepackage{url}            
\usepackage{booktabs}       
\usepackage{amsfonts}       
\usepackage{nicefrac}       
\usepackage{microtype}      
\usepackage{xcolor}         

\usepackage{algorithm}
\usepackage{algorithmic}

\usepackage{graphicx}
\usepackage{subcaption}
\usepackage{multirow}
\usepackage{amsmath}
\usepackage{amsfonts}
\usepackage{bm}
\usepackage{dirtytalk}
\usepackage{comment}
\usepackage{wrapfig}
\usepackage{pifont}
\newcommand{\cmark}{\ding{51}}%
\newcommand{\xmark}{\ding{55}}%
\usepackage[toc,page]{appendix}

\usepackage{hyperref}       
\definecolor{mydarkblue}{rgb}{0,0.08,0.45}
\hypersetup{colorlinks,linkcolor=red,urlcolor=mydarkblue,linktoc=page}

\title{General Characterization of Agents by\\States they Visit}

\author{%
  Anssi Kanervisto\\
  University of Eastern Finland\\
  \texttt{anssk@cs.uef.fi}\\
\And
  Tomi Kinnunen\\
  University of Eastern Finland\\
\And
  Ville Hautam\"aki\\
  University of Eastern Finland\\
  National University of Singapore
}

\begin{document}

\maketitle

\begin{abstract}
  \textit{Behavioural characterizations} (BCs) of decision-making agents, or their policies, are used to study outcomes of training algorithms and as part of the algorithms themselves to encourage unique policies, match expert policy or restrict changes to policy per update. However, previously presented solutions are not applicable in general, either due to lack of expressive power, computational constraint or constraints on the policy or environment. Furthermore, many BCs rely on the actions of policies. We discuss and demonstrate how these BCs can be misleading, especially in stochastic environments, and propose a novel solution based on what states policies visit. We run experiments to evaluate the quality of the proposed BC against baselines and evaluate their use in studying training algorithms, novelty search and trust-region policy optimization. The code is available at \url{https://github.com/miffyli/policy-supervectors}.
\end{abstract}

\section{Introduction}
    
    
    


    While creating or training autonomous agents, whether it is via manual coding, \textit{reinforcement learning} (RL) \cite{sutton2018reinforcement} or \textit{evolution strategy} (ES) \cite{hansen2001completely} algorithms, one wishes to compare the solutions to find out which work and which do not. A common approach is comparing the performance of agents' policies in the given task \cite{henderson2018deep} or by studying how policy behaves \cite{pitis2020maximum}. In addition to these explicit comparisons, RL and ES algorithms include implicit comparisons to encourage finding novel solutions \cite{lehman2008exploiting, conti2018improving, eysenbach2018diversity, parker2020effective}, to match the behaviour of an expert policy \cite{ho2016generative, ni2020f} or to limit changes on policy's behaviour to avoid catastrophic failure \cite{schulman2015trust, ppo}.
    
    To facilitate such comparisons, \textit{behavioural characterizations} (BCs)~\cite{lehman2008exploiting} aim to capture policy's behaviour in a fixed representation accompanied by a distance metric. A general BC could be used to generalize previously proposed methods \cite{conti2018improving, pacchiano2020learning}. However, BCs in the previous work are often domain-specific or depend on policy's actions which are not descriptive of behaviour in general, as we will discuss in Section \ref{sec:choosing-bc}.

    Recent work attempted to generalize the notion of a \textit{behavioural embedding} \cite{pacchiano2020learning}, considering policies as distributions over trajectories, and providing a principled mechanism to compare policies in the behavioural space. However, this approach relies on \say{behavioural embedding mappings}, which map trajectories to structures that are believed to describe the trajectory (e.g., reward-to-go, terminal state). It remains unclear how the choice of this mapping affects the results.
    
    In this work, we propose and evaluate a more general BC that can be applied to different domains. Our work provides the following contributions: (1) We summarize and compare BCs used in the previous work. (2) We discuss and demonstrate the shortcomings of action-based BCs which have been actively used in the previous work. (3) Propose a novel BC method based on which states the agent visits, borrowing techniques from the field of speaker recognition~\cite{kinnunen2010overview}. (4) We use the proposed BC to study and visualize different training algorithms and also explore their use in novelty search \cite{lehman2008exploiting} and trust-region policy optimization \cite{schulman2015trust}.

\section{Preliminaries and definitions}
    We model environments as stochastic \textit{Markov decision processes} (MDPs) \cite{sutton2018reinforcement}, where agent acts on states $s \in \mathcal S$. The environment begins in an initial state $s_0 \sim p(s_0)$. The policy $\pi \in \Pi$ provides an action $a \in \mathcal A$ as a stochastic function $a \sim \pi(s)$. After executing an action, the environment evolves to the next state according to a stochastic function $s' \sim p(s' | s, a)$ (transition dynamics), and agent is provided with a reward $r \sim p(r | s, a, s')$. This process repeats until environment lands into a terminal state $s_T \in \mathcal S$. A single episode begins from an initial states and ends to a terminal state. A trajectory is a tuple of states and actions experienced during one episode. An environment is defined by tuple $(\mathcal S, \mathcal A, p(s_0), p(s' | s, a), p(r | s, a, s'), \mathcal T)$.
    
    Behaviour characterization is a function $b: \Pi \rightarrow \mathcal B$, where $\mathcal B$ is space of all behaviours, accompanied with a \textit{pseudo-distance} function $d: \mathcal B \times \mathcal B \rightarrow \mathbb R^+$. This need not be a proper distance metric, as long it is non-negative and symmetric. Ideally, the more different behaviour of two policies is, the larger this distance should be. Note that $b$ may be a stochastic function if, for example, it relies on randomly sampled data to map policy to a behaviour.

\section{Choosing behavioural characterization}
\label{sec:choosing-bc}
    \begin{table}[]
        \caption{Comparison of different BCs. Required compute describes CPU and memory requirements for comparing a large set of policies, excluding the sampling of the environment. Environment agnostic methods do not pose requirements for the environment, and action agnostic methods can compare policies across action spaces.}
        \label{tab:policy-comparisons}
        \centering
        \begin{tabular}{lcccccc}
        \toprule
\multirow{2}{*}{\textbf{Description and reference}} & \multirow{2}{*}{\textbf{\begin{tabular}[c]{@{}c@{}}Required\\compute\end{tabular}}}  & \multirow{2}{*}{\textbf{Describes}}   & \multirow{2}{*}{\textbf{Expressiviness}}   & \multicolumn{2}{c}{\textbf{Agnostic}}                \\
\cline{5-6}  &   &   &   & \textbf{Env.} & \textbf{Action} \\  \midrule
Returns \cite{henderson2018deep}              & Low                                                     & Policy                    & Low                     & \cmark                   & \cmark               \\
Policy parameters \cite{gaier2019weight}                   & Low                                                & Policy                    & Low                     & \cmark               & \xmark                       \\
Comparing actions \cite{pacchiano2020learning, harb2020policy}          & Low                                                     & Policy                    & Low                     & \cmark               & \xmark                       \\
Termination state \cite{conti2018improving}                    & Low                                                     & Policy                    & Low                     & \xmark               & \cmark               \\
Transition matrix \cite{matusch2020evaluating} & V.High                                    & Policy                    & High                    & \xmark              & \xmark                   \\
Trajectories \cite{conti2018improving}  & High                                                    & Trajectory                & High                    & \xmark                 & \cmark               \\
States + Discriminator \cite{eysenbach2018diversity}     & N/A                                                     & Skill \cite{eysenbach2018diversity}                    & High                    & \cmark               & \cmark               \\
States + Gaussian \cite{berseth2019smirl}          & Low                                                     & Policy                    & Low                     & \cmark               & \cmark               \\ 
Trajectory encoder \cite{wang2017robust, raileanu2020fast}                   & High                                                     & Trajectory                & High                    & \cmark              & \xmark                   \\
Aggregate state-actions \cite{grover2018learning, tang2020represent}                   & High                                                     & Policy                & High                    & \cmark              & \xmark                   \\ \midrule
States + Discriminator (Adapted)     & Low                                                    & Policy                    & High                    & \cmark               & \cmark              \\ 
Trajectory encoder (Adapted)                   & High                                                     & Policy                    & High                    & \cmark               & \cmark               \\ \midrule
States + GMM (Proposed)                  & Low                                                     & Policy                    & High                    & \cmark               & \cmark               \\ \bottomrule
\end{tabular}
    \end{table}
    
    Table \ref{tab:policy-comparisons} compares various BCs used in the previous research, which have been either used explicitly for visualization (returns, terminal state) or implicitly in the training algorithm (e.g., novelty search \cite{stanley2002evolving}, diversity \cite{eysenbach2018diversity}). Table \ref{tab:policy-comparisons} also includes methods which model trajectories or skills (e.g., fixed policy conditioned on a latent vector), rather than policies.
    
    While we could use policy's parameters to compare agents, this does not generalize over different types of algorithms nor can we tell if a change in parameter vector is truly meaningful to the agent's behaviour \cite{conti2018improving, harb2020policy}. Episodic returns are the \textit{de facto} approach for measuring quality of policies, but policies with distinct behaviour can achieve same returns (see Figure \ref{fig:gridworld}) \cite{agarwal2021pse}. Using a termination state, such as the final coordinates of the robot, has been successful in novelty search but is limited to environments where such heuristics can be used. Element-wise comparisons of trajectories require either fixed-length episodes or heuristics to combine varying length episodes. Finally, one can estimate transition dynamics by sampling the environments~\cite{matusch2020evaluating}, but this has a high computational cost that limits the number of policies compared. With these options discarded, we are left with BCs which focus on actions policy takes or states they visit.
    
    \subsection{Actions and stochastic environments}
        \label{sec:actions-and-stochastic}
        \begin{wrapfigure}[21]{right}{0.5\textwidth}
            \centering
            \includegraphics[width=0.5\textwidth]{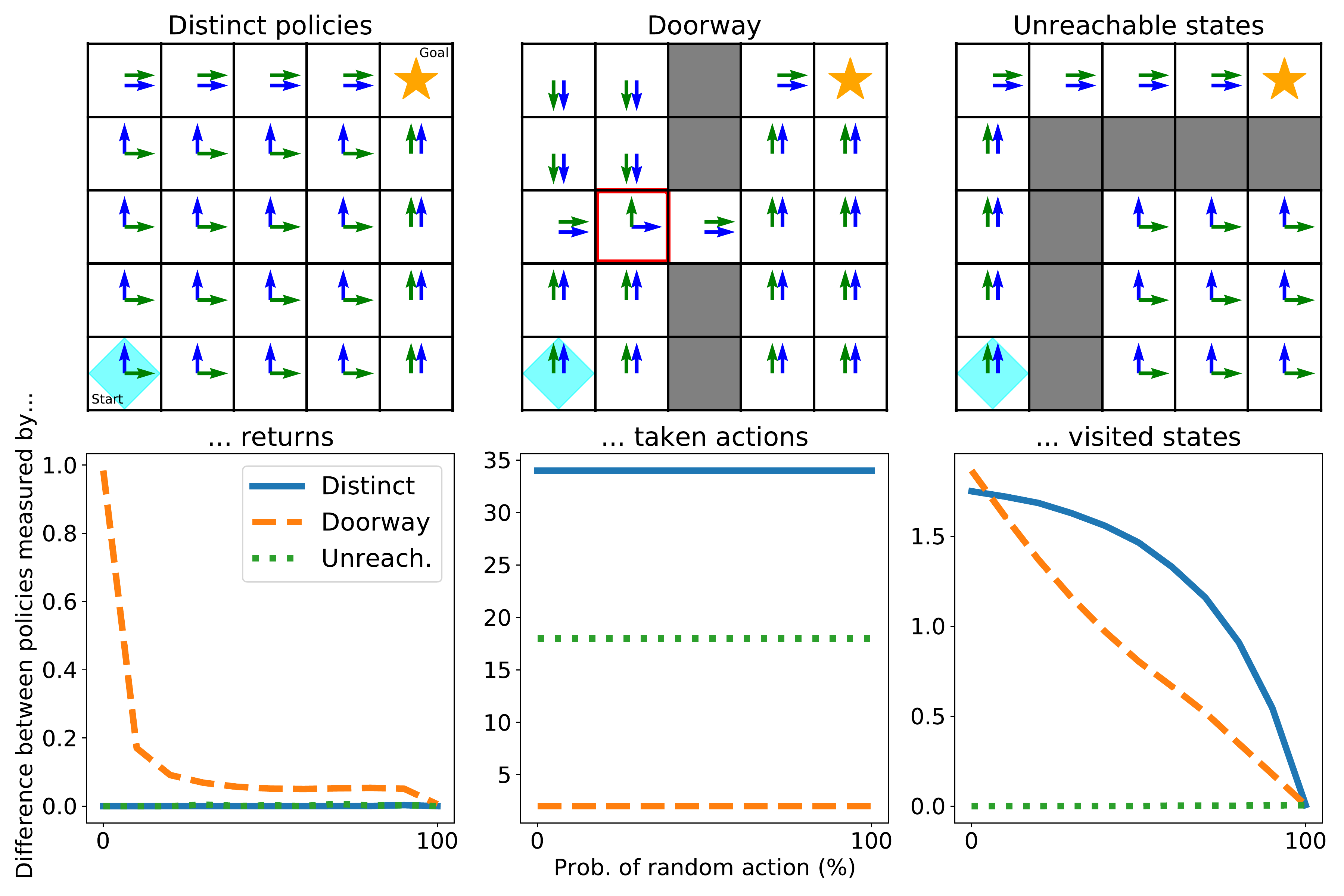}
            \caption{Difference between the two policies, blue and green, as measured by different BCs. State BC is obtained with ten thousand trajectories per policy, and distance is measured as the sum of the absolute errors over states. Action BC difference is the sum of the absolute difference between action distributions. Details are available in Appendix~\ref{sec:appendix-gridworld}.}
            \label{fig:gridworld}
        \end{wrapfigure}
        Comparing policies by what actions they chose is intuitively a sound solution, as one may argue the policy's actions are what define its behaviour. Action-based BCs are used in gradient-based RL algorithms to restrict changes to policy's behaviour \cite{schulman2015trust, ppo} or to encourage diversity \cite{parker2020effective}. This is done by comparing actions of policies they would take in a fixed set of states (an \say{off-policy embedding} \cite{pacchiano2020learning}).

        However, using actions alone for BC would ignore the transition dynamics of the environment. As discussed by Pacchiano et al.~\cite{pacchiano2020learning}, a small change to actions could lead the agent to wildly different states or a large change in actions could have no effect.\footnote{If an agent takes an action and it has no effect on the environment, did the agent take an action from an outside observer's point of view?} To illustrate this, Figure~\ref{fig:gridworld} shows three scenarios where the behaviour of two policies are represented with the average episodic return, distribution of the taken actions and distribution of the visited states.
        
        Action-based BC is agnostic to environment stochasticity. As stochasticity of the environment increases, the policies tend towards random behaviour, and action-based BC describes them as two very different policies while they both behave the same (random agents). Another weakness is the insensitivity to \say{doorway} scenarios, where a single action can lead to different states. Action-based BC shows the two policies are very similar, but the green policy never reaches the other side of the world until we increase stochasticity. In real environments, doorway scenarios may manifest as literal doorways, critical points in grasping an object or pressing a button in a video game. Finally, even if some of the states used to compare action-differences were unreachable, they affect the results. In the right-most figure of Figure~\ref{fig:gridworld}, both policies behave exactly the same in the area where they can traverse, yet the action-based distance is high. 
        
        In summary, while action-based BCs are useful for RL training, they can not be relied upon as a general BC: without accounting for the environment dynamics, one can not say if the difference in actions is meaningful.
    
    \subsection{Describing policies by states they visit}
        A common alternative \cite{eysenbach2018diversity, berseth2019smirl} to action-based BCs is comparing \textit{what states policies visit}. This requires sampling states for each policy, but in turn, it captures the environment dynamics. Referring to Figure \ref{fig:gridworld}, where action-based BC provided misleading descriptions, state-based BC captures correctly the increasing stochasticity and the effect of the doorway and unreachable states. Compared to returns, it also captures the difference between two optimal but distinct policies. 
        
        A simple state-based BC is to fit a multivariate Gaussian on the states visited by the policy~ \cite{berseth2019smirl}. Both fitting and comparing policies is fast (e.g., KL-divergence, which can be evaluated in a closed-form), but limits the description to a unimodal distribution. The true distribution of states can be complex even for a simple environment (see scatter plots in Figure \ref{fig:pivector-extraction}). Instead of a single Gaussian, a neural network discriminator can be trained to measure the probability of a state coming from a given policy, as done by Eysenbach et al.~\cite{eysenbach2018diversity} and Ni et al.~\cite{ni2020f}, Alternatively, one can train an auto-encoder to encode varying-length trajectories into fixed length~\cite{wang2017robust}. Alas, the BCs of these works are tightly integrated with the training loop, and can not be used with policies trained by other algorithm. Instead, we propose using a mixture of Gaussian models to model complex distributions of states, which is detailed in the next section.

\section{Policy supervectors}
    \label{sec:supervector}
    We consider a setup with $N$ policies $\pi_i, \; i \in \{1 \ldots N\}$ which we want to compare. For each policy, we play $M$ episodes and store all encountered states $\bm s$ to a per-policy dataset $\bm B_i = (\bm s_{\{i,1\}}, \bm s_{\{i,2\}}, \ldots)$. Note that the stored states may differ from states the policy acts upon; stored states can be a separate piece of information believed to describe the behaviour, such as x-y coordinates of a maze robot \cite{pitis2020maximum}. In the case of high-dimensional environments, one can train an encoder to turn images into more compact latent codes \cite{berseth2019smirl} or to preprocess them into low-dimensional representations \cite{ecoffet2019go}.
    
    \begin{wrapfigure}[26]{r}{0.4\textwidth}
        \centering
        \includegraphics[width=0.4\textwidth]{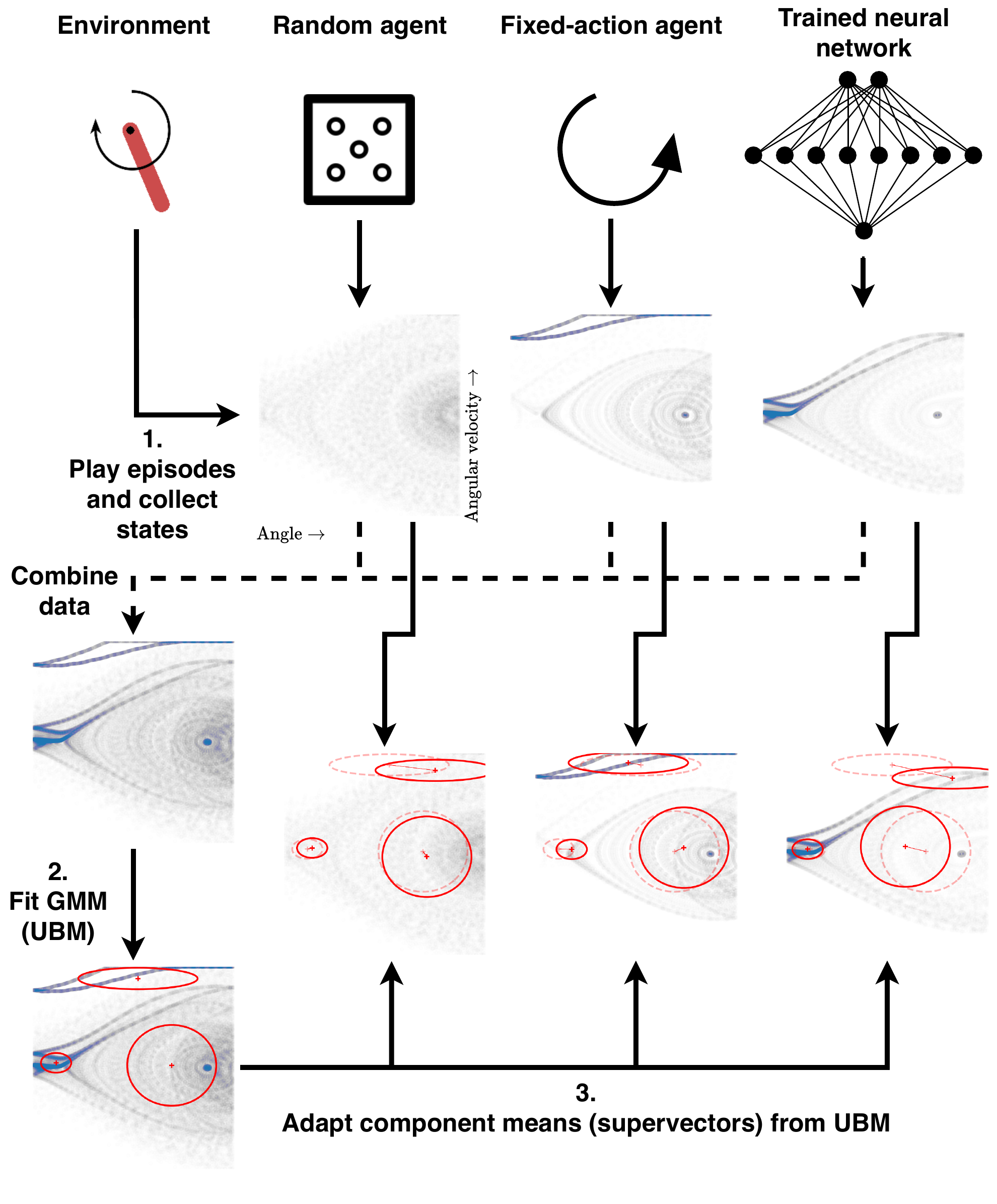}
        \caption{
        Computing policy supervectors for three policies in the Pendulum environment, using real data and policies. Each faint dot represents a single state of the Pendulum environment, and red circles represent Gaussian components of the GMMs.
        }
      \label{fig:pivector-extraction}
    \end{wrapfigure}
    
    \textit{Gaussian mixture models} (GMMs) \cite{mclachlan1988mixture} can be used to model multi-modal data, and with enough components they can model almost any continuous distribution to an arbitrary accuracy \cite{bishop2006pattern}. However, measuring KL-divergence requires expensive sampling. While approximations exist, they come with different assumptions and drawbacks \cite{hershey2007approximating}.
    
    Instead, we draw inspiration from the field of speaker verification where GMMs have been used to model speaker characteristics \cite{kinnunen2010overview}. Instead of training one GMM per speaker, Reynolds et al.~\cite{reynolds2000speaker} presented \textit{universal background models} (GMM-UBMs, later UBMs) to model a distribution of speaker-independent features by fitting a large GMM on a large pool of speaker data. Speaker-specific model can then be adapted with \textit{Maximum a Posteriori} (MAP) adaptation of GMM components~\cite{gauvain1994maximum} using individual speaker's data. The acquired parameters can then be concatenated into a long vector, called a \textit{supervector}~\cite{campbell2006support}, describing the speaker with a fixed-length vector. 
    
    We apply this method to policies instead of speakers. We first fit a GMM on state datasets from all policies and then MAP-adapt a supervector for each policy, which we call a \textit{policy supervector}. Under the Bayesian view, the UBM represents an informative prior while the adapted GMM is a point estimate of the posterior, where we use the same data as part of the prior and adapting the posterior. We treat UBM as a common feature extractor, and by including data from all policies we ensure the prior covers all states visited by all of the policies. This process is illustrated in Figure \ref{fig:pivector-extraction}. 

    Formally, policies' datasets are pooled together into an UBM-training set $\bm B_\text{ubm} = (\bm B_1, \ldots , \bm B_N)$. This set is then used to fit an UBM with $K$ components and parameters $\bm \mu_\text{ubm} \in \mathbb R^{K \times d}$, $\bm \Sigma_\text{ubm} \in \mathbb R^{K \times d \times d}$ and $\bm w_\text{ubm} \in \mathbb R^K, w_k \geq 0, \sum_k w_k = 1$ with EM-algorithm until convergence. We can then perform MAP-adaptation on a per-policy dataset $\bm B$ to obtain adapted mean $\bm{\hat \mu}_k$ of $k$\textsuperscript{th} component with \cite{reynolds2000speaker}
    \begin{align}
        p(k | \bm s_t) =& \frac{\bm w_k \mathcal N(\bm \mu_k, \bm \Sigma_k)}{\sum^K_{l=1} \bm w_l \mathcal N(\bm \mu_l, \bm \Sigma_l)} \label{eq:resp}\\
        n_k =& \sum_{t=1} p(k | \bm s_t)\\
        E_k(\bm B) =& \frac{\sum_{t=1} p(k | \bm s_t) \bm s_t}{n_k} \\
        \alpha_k =& \frac{n_k}{n_k + r} \\ 
        \bm{\hat \mu}_k =& \alpha_k E_k(\bm B) + (1 - \alpha_k) \bm \mu_k \label{eq:map-means},
    \end{align}
    where $r$, known as a \textit{relevance factor}, is a control parameter that impacts how much new data can affect the adapted mean.
    
    To measure the distance between two policy supervectors, Campbell et al.~\cite{campbell2006support} show that an upper-bound for KL-divergence of means of two adapted GMMs $\bm \mu_i$ and $\bm \mu_j$ is
    \begin{equation}
        \label{eq:distance}
        d_{\text{KL}}(\bm \mu_i, \bm \mu_j) \leq \frac{1}{2} \sum^K_{k=1} w_\text{ubm}^k (\bm \mu_i^k - \bm \mu_j^k) \bm \Sigma_\text{ubm}^{-1} (\bm \mu_i^k - \bm \mu_j^k) \; .
    \end{equation}
    Unlike KL-divergence, this upper-bound is symmetric \cite{campbell2006support}. In this work, we will only adapt the means of the UBM to allow the use of the above metric, but this method can be extended by adapting covariances and weights (see \cite{reynolds2000speaker} equations (11)-(13)). We use this upper bound as a distance metric for policies. 
    
    \textbf{Limitations.} Policy supervectors require each policy to sample data from the environment which can be expensive in slow environments. In highly stochastic environments we need more samples to accurately describe policies, which will make the training phase of these methods more demanding. Policy supervectors also require the environment to be the same for all policies, which can be difficult for real-world robotics, for example.

\section{Adapted baselines from related work}
    \label{sec:adapted-baselines}
    Given the number of work towards characterising policies (Table \ref{tab:policy-comparisons}), we adapt two of the previous work to form two baseline solutions. These methods share the same limitations as supervectors discussed above.
    
    \subsection{Discriminator as a state-density estimator}
        \label{sec:discriminator}
        Motivated by the earlier use of discriminator networks as a density estimator~\cite{eysenbach2018diversity, ni2020f}, we train one network per policy to distinguish between states this policy encounters versus all other policies. Specifically, we train a discriminator neural network $D_\pi$ by ascending the loss
        \begin{equation}
            \mathcal L_i = \mathbb E_{\bm s \sim \bm B_i} \left [ \log D_i(\bm s) \right ] + \mathbb E_{\bm s \sim \bm B_{j \neq i} } \left [ \log (1 - D_i(\bm s)) \right ] ~ ,
        \end{equation}
        where $i$ is the index of the policy we are about to compare to others. Optimal discriminator satisfies properties \cite{goodfellow2014generative, ni2020f}
        \begin{align}
            D^*(\bm s) &= \frac{p_i(\bm s)}{p_i(\bm s) + p_{j \neq i}(\bm s)} \\
            \Rightarrow \frac{p_i(\bm s)}{p_{j \neq i}(\bm s)} &= \frac{D^*(\bm s)}{1 - D^*(\bm s)} = r_i(\bm s) ~ ,
        \end{align}
        where $p_i(\bm s)$ is shorthand for $p_{\bm B_i}(\bm s)$. This value $r_i(\bm s)$ represents how likely it is the state was observed by $\pi_i$ than by the rest of the policies. With this in mind, we define a pseudo-distance between two policies,
        \begin{align}
            f_i(\bm s) &= e^{-\log r_i(\bm s)} \label{eq:discriminator-distance}\\
            d(i, j) &= \mathbb E_{\bm s \sim \bm B_i} \left [ f_j(\bm s) \right ] + \mathbb E_{s \sim \bm B_j} \left [ f_i(\bm s) \right ] ,
        \end{align}
        summing distance both ways for symmetry. Essentially, for each policy, we measure how likely it is that their data is also contained in others' datasets and vice versa. In practice, the expectations are computed as averages over the datasets. We opt for training one discriminator per policy (policy's states versus all others) rather than one per comparison (policy X's states versus policy Y's) to avoid the quadratic explosion of discriminator training operations required. Training details can be found in Appendix~\ref{sec:appendix-discriminator}.
    
    \subsection{Trajectory encoder}
        \label{sec:encoder}
        Wang et al.~\cite{wang2017robust} proposed to train a \textit{variational autoencoder} (VAE)~\cite{kingma2013auto} to reconstruct trajectories. The encoder turns a sequence of states (trajectory) into an embedding $\bm z$, and the decoder reconstructs original state-action pairs from this embedding. We adapt this method for policies by modelling the distribution of trajectory embeddings. We train the system on data from all policies, then use the encoder to sample one embedding per trajectory, and then fit a multivariate Gaussian on these embeddings. The distance between two policies is then symmetric KL-divergence of these two Gaussians, where we sum the KL-divergences both ways. Further details are available in Appendix~\ref{sec:appendix-encoder}.

        Compared to the two other proposed methods, this method can capture temporal dependencies between states, while the other two methods discard this information. However, this could lead to posterior collapse where the decoder predicts future states purely based on previous states, ignoring the embedding. If the environment and/or policy are complex, we need a complex model to encode the whole policy into a fixed embedding. Finally, out of the three methods, training the encoder is the most computationally expensive.

\section{Related work}
    The \textit{de facto} method for comparing policies is by evaluating their performance in a task \cite{jordan2020evaluating}, or if the environment permits it, we can compare the different coordinates policy visits \cite{pitis2020maximum, conti2018improving}. Hernandez et al. ~\cite{hernandez2020comparison} make use of dimensional reduction to plot fixed-length trajectories of different self-play algorithms to study the evolution of policies, and Matusch et al.~\cite{matusch2020evaluating} compare discretized transition matrices to study which metrics correlate with human behaviour. Others propose decompressing compact representations of policies back into functional policies, such as from genomes \cite{gaier2019weight} or random generator seeds \cite{such2017deep}.
    
    Trust-region policy optimization \cite{schulman2015trust} limits per-update change to policy's behaviour to ensure improvement in returns, but similar methods can also be used to encourage novel behaviour \cite{lehman2008exploiting, conti2018improving} or diverse behaviour \cite{eysenbach2018diversity, parker2020effective}. An RL agent can be encouraged to visit unseen states \cite{bellemare2016unifying}, where visit counts can be estimated by prediction error \cite{burda2018exploration}. Framing imitation learning as a task of matching state or state-action distributions has also been successful \cite{ho2016generative, ni2020f, levine2020offline}. Compact policy embeddings have also been used to create generalized value functions~\cite{raileanu2020fast, tang2020represent, harb2020policy} or to improve imitation learning~\cite{wang2017robust}.
        
\section{Experiments and results}
    For an empirical assessment of the proposed BC (policy supervectors), we first evaluate it against the baseline BCs and then explore their use in various tasks. We aim to answer the following research questions (RQs). \textbf{RQ1} Does the BC separate truly different policies (expressivity)? \textbf{RQ2} How sensitive are the BCs to the random sampling of the environment? If BC does not describe the policy, there is no reason to use it. If it produces wildly different results even with a lot of data, its results may not be trusted. Note that we have not included action-based BCs because of the reasons discussed in Section \ref{sec:actions-and-stochastic} (their measurements can not be trusted). However, we will compare to them in trust-region experiments later (Section \ref{sec:trust-region}).
    
    We use five classic control environments from the OpenAI Gym library \cite{gym} which include low dimensional environments (Cartpole, Pendulum and Acrobot) as well as two higher-dimensional environments (Lunarlander and Bipedal-walker). We opt for these simple environments to allow scaling the experiments. We train three \textit{proximal policy optimization} (PPO) \cite{ppo} agents per environment and store $100$ versions of the policy during training. For each policy, we then collect a set of trajectories, which we use to compute the state-based BCs and finally measure distances between these $100$ policies. We repeat this data collection and distance measurement three times. Distance matrices are min-max normalized to $[0, 1]$ to allow comparison between different repetitions. Technical details, source code and further results described can be found in the Appendix~\ref{sec:appendix-evaluation}.

    \subsection{Baselines}
        In addition to adapted baselines (Section \ref{sec:adapted-baselines}), we include two baseline solutions.

        \textbf{Discretization.} Similar to Matush et al.~\cite{matusch2020evaluating}, we discretize each state dimension to ten equally spaced bins, count occurrences of each bin and divide by the number of states visited. The distance between two policies is then $d(p_i, p_j) = \frac{1}{2} \sum_s | p_i(s) - p_j(s) |$. We opt to use this sum over KL-divergence due to the prevalence of zeros in the distributions. While simple, this method's computational requirements explode with the increasing number of dimensions.

        \textbf{Single Gaussian.} We fit a multivariate Gaussian on the sampled states and define distance between two policies as $d_{\text{KL}}(p_i || p_j) + d_{\text{KL}}(p_j || p_i)$ for symmetry. This has been used previously in tracking what states policy has visited \cite{berseth2019smirl}.
    
    \subsection{Evaluation metrics}
        \label{sec:evaluation-metrics}
        \begin{wrapfigure}[35]{r}{0.6\textwidth}
            \centering
            \includegraphics[width=0.6\textwidth]{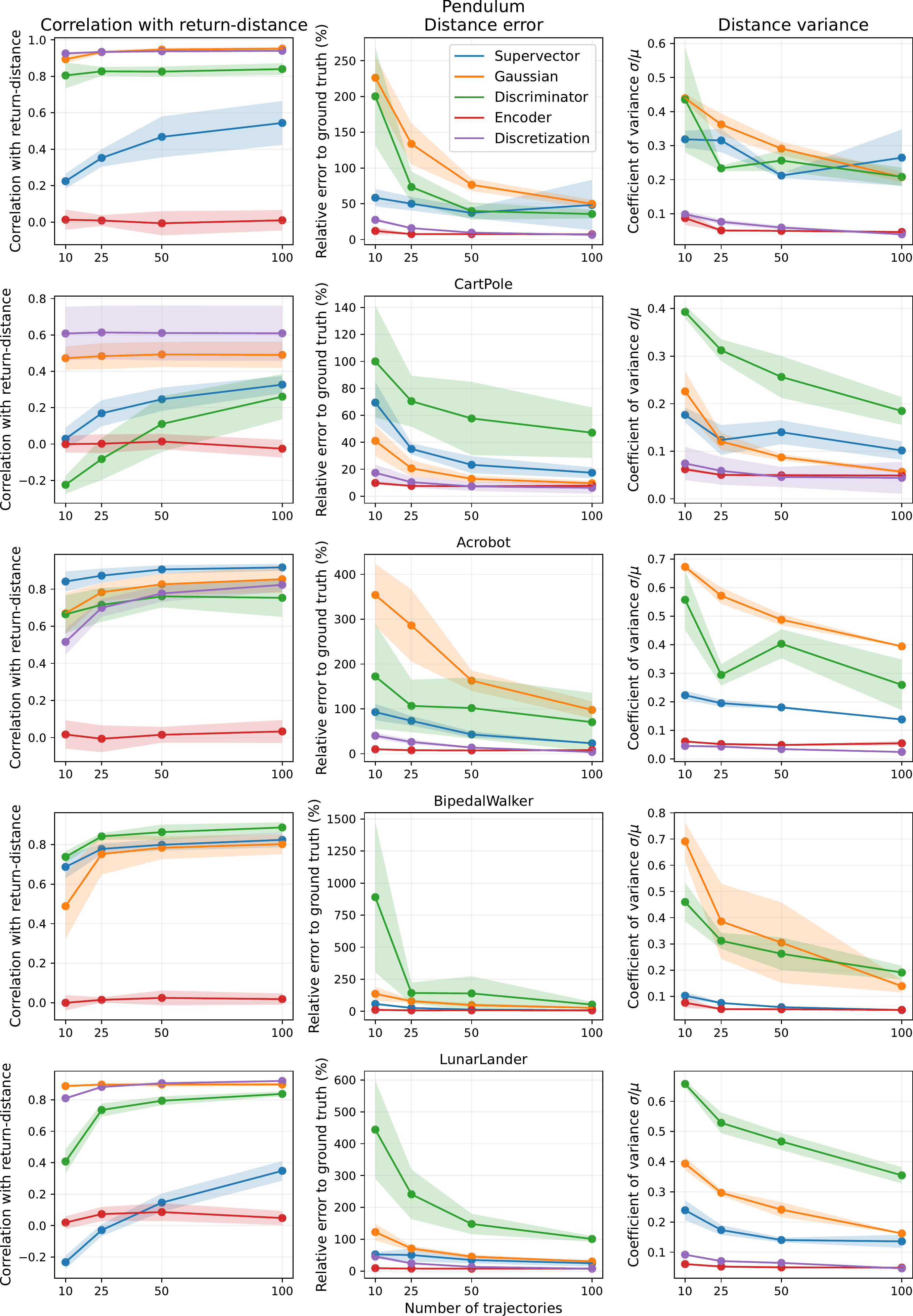}
            \caption{Evaluation results of different BCs. Averaged over three repetitions. The shaded region is plus/minus one standard deviation.
            }
            \label{fig:metric_results}
        \end{wrapfigure}
        \textbf{Correlation with return-distance.} To measure how well BC separates different policies, we measure the Pearson correlation between the absolute difference between average returns and distances measured by the BC. While returns are limited in their expressiveness, it still indicates the policies are doing something different if the returns differ. This especially applies to the environments we use, as the reward signal is tied to the states agent visits.
        
        \textbf{Distance error.} Given a \say{ground truth} distance between policies under the same BC, we measure the average relative error between this ground truth and predicted distances. We select one of the repetitions with the highest amount of data ($100$ trajectories) to represent this ground truth. This metric aims to measure how sensitive the method is to the amount of data we collect.
        
        \textbf{Distance variance.} As behavioural distances between a fixed set of policies should stay the same relative to each other, we study the variance in the results by computing \textit{coefficient of variation} (CV) \cite{everitt2002cambridge} $\sigma / \mu$ over the repetitions, where $\sigma$ is the standard deviation of the distance over repetitions and $\mu$ is the sample average of distances. A lower value indicates more similar results over repetitions.

    \subsection{Evaluation results}
        \label{sec:main-results}
        Figure \ref{fig:metric_results} shows the results for each environment separately. The single Gaussian and discriminator methods correlate with return differences, but require many trajectories per policy to stabilize results. Policy supervectors with $64$ Gaussian components offer stabler results even at a lower number of trajectories, which we believe is due to the use of data from all policies in the UBM training. By sweeping over the different number of components, we find a strong connection between environments and the optimal number of components: in some environments one to four components provides the stablest results (Appendix~\ref{sec:appendix-evaluation}).

        The discretization method appears to capture the behaviour as well as remain stable, but this method is computationally limited: experiments in the BipedalWalker environment ran out of memory (64GB of system memory) when trying to build the transition matrix over multiple dimensions. Trajectory encoders were similarly limited, which we found to be slow to train due to long sequences. We believe the low performance of encoder method is due to the difficulty of predicting future states. In summary, we find policy supervectors a scalable and functioning state-based BC, with discretization a viable option in low-dimensional environments but not generally applicable due to memory constraints.

    \subsection{Studying evolution of policies under training algorithms}
        \label{sec:evolution-of-policies}
        \begin{figure}[t]
            \centering
            \includegraphics[width=1.00\columnwidth]{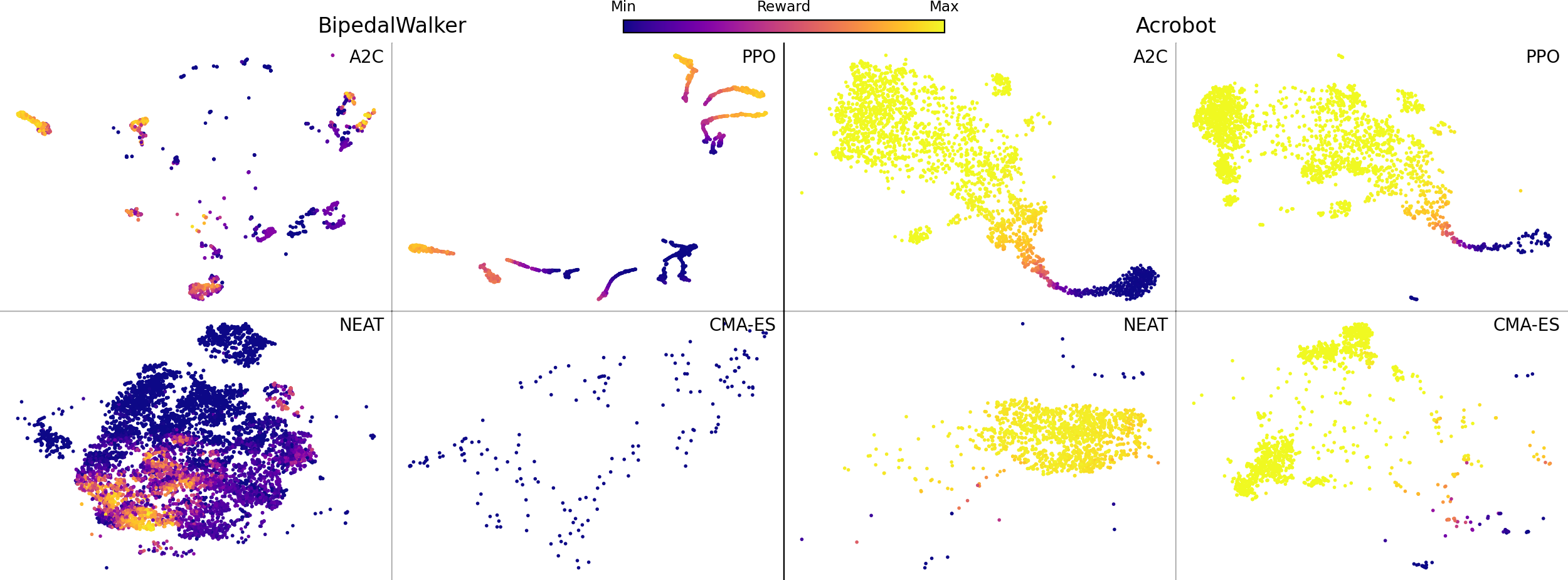}
            \caption{t-SNE plot of the evolution of policies under different training algorithms, where each point represents a single policy. Plots under the same environment share the same plot scales. Rewards are scaled according to minimal and maximal attainable reward per environment.}
            \label{fig:tsne-overview}
        \end{figure}
        Evolution-based methods such as \textit{neuroevolution of augmenting topologies} (NEAT) \cite{stanley2002evolving} and \textit{covariance matrix adaptation ES} (CMA-ES) explore by modifying promising solutions and testing which location of the parameter space works better \cite{hansen2001completely}, while gradient-based RL methods like \textit{advantage actor critic} (A2C) \cite{a3c} and PPO update a single policy towards higher episodic reward. To better understand how these training algorithms evolve policies, we train agents with them and store checkpoints during training. We then extract policy supervectors of these checkpoints (with $64$ components) and plot the resulting points with t-SNE \cite{maaten2008visualizing} dimensionality reduction using the adapted distance \eqref{eq:distance}. The hypothesis is that evolution-based methods cover a wider area of policies (random mutation of the parameters leads to different behaviours), while RL algorithms remain in a small region per run. We are not comparing which of these algorithms is better, rather we aim to understand if they explore different behaviours and how they evolve the policies.
        
        Figure \ref{fig:tsne-overview} shows the results with five A2C/PPO runs and one NEAT/CMA-ES run. In total, roughly 30,000 policies are compared against each other, with a varying number of policies from different algorithms depending on the settings (see Appendix~\ref{sec:appendix-algo-evaluation} for details and remaining plots). NEAT and CMA-ES cover a wide area of behaviours, as expected.
        RL solutions cover smaller areas, with PPO forming distinguishable \say{worms} in more complicated environments (BipedalWalker), while A2C forms small clouds. This suggests the trust-region restriction of PPO is visible in policy supervectors as well, where the change in behaviour between successive policies is small, whereas A2C updates may change behaviour considerably.

        Indeed, we find a positive correlation between total distance travelled by the trained policy and PPO ratio-clip value (Appendix \ref{sec:appendix-algo-evaluation}), which controls the size of the trust region. We also find a negative correlation between distance travelled after an update and average return, indicating that the initial learning steps change the agent's behaviour the most. These results concur with observations of Engstrom et al.~\cite{engstrom2020implementation}, where the KL-distance between successive policies first increases and then decreases as training moves on. The results with \textit{behavioural cloning} \cite{pomerleau1989alvinn} indicate that distance between the expert policy and trained policy decreases as training progresses, with change to behaviour slowing down as training reaches closer to the expert's performance.
    
    \subsection{Applying supervectors to trust-region policy optimization}
        \label{sec:trust-region}
        
        
        
        

        
        
        The above insights on trust-region optimization and BC distances suggest that state-based BCs could also be used in trust-region optimization. To test this, we construct a $N$-dimensional grid world environment where the agent can move to one of $N$ directions per grid, where one direction is correct ($+1$ reward) and rest either reset the episode or do nothing. This environment is constructed to reflect the doorway scenario: some actions have a large impact, while others do not. Agents are trained with a modified PPO where we disable the policy ratio clip and instead check if state-based BC distance is larger than the threshold after every network update. We compare state-based BCs against no constraint and \textit{total variation divergence} (Max TV) $d(\pi, \pi') = \max_s \frac{1}{2} \sum_a |\pi(a|s) - \pi'(a|s)|$~\cite{schulman2015trust}. We sweep over threshold values and present results for the method with the largest area under the learning curve. Further details are available in Appendix~\ref{sec:appendix-trust}.
        
\begin{wrapfigure}{r}{0.5\textwidth}
            \centering
            \begin{subfigure}[b]{0.24\columnwidth}
                \centering
                \includegraphics[width=\columnwidth]{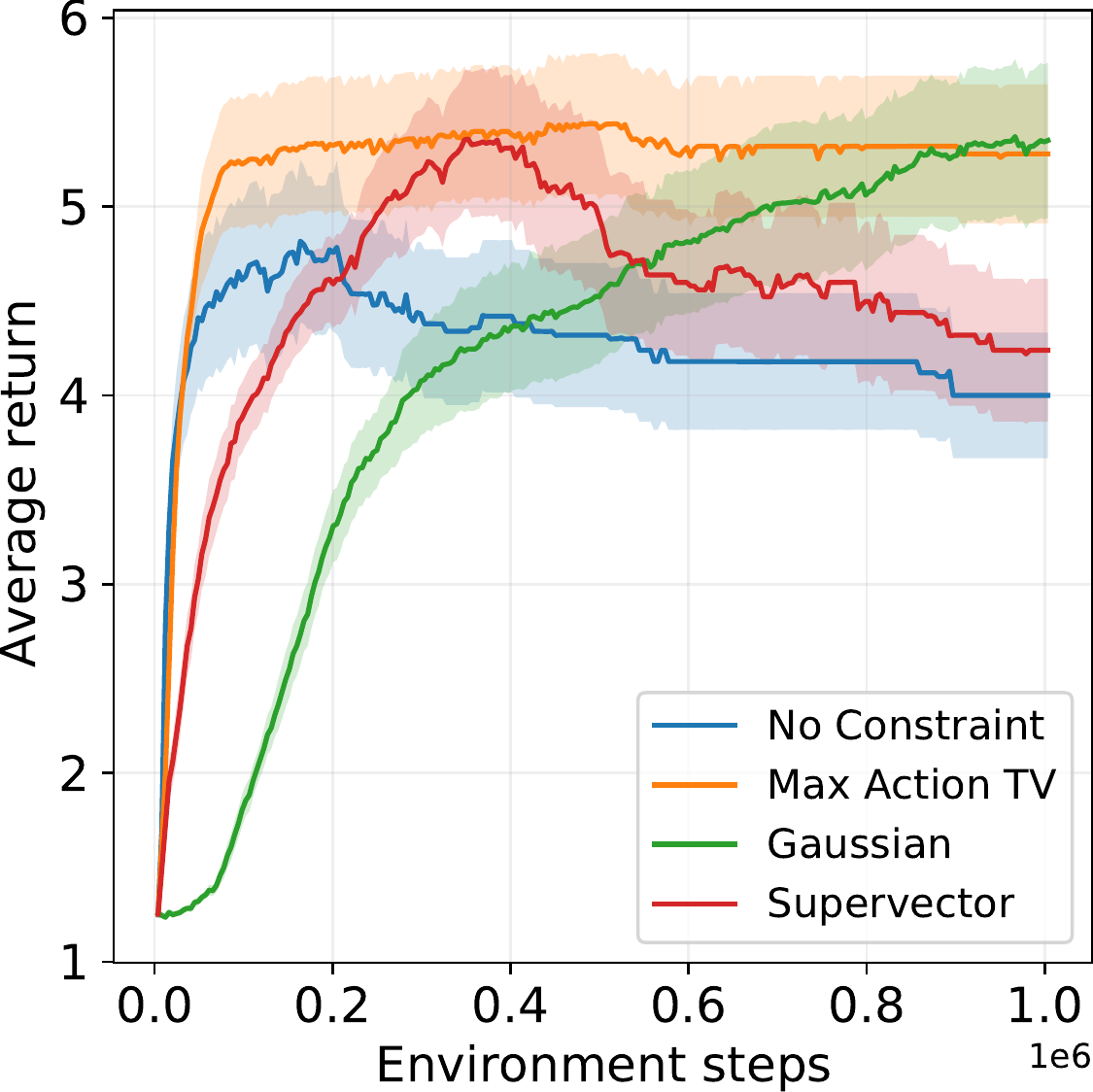}
                \caption{Trust region}
                \label{fig:trust-region-results}
            \end{subfigure}
            \begin{subfigure}[b]{0.24\columnwidth}
                \centering
                \includegraphics[width=\columnwidth]{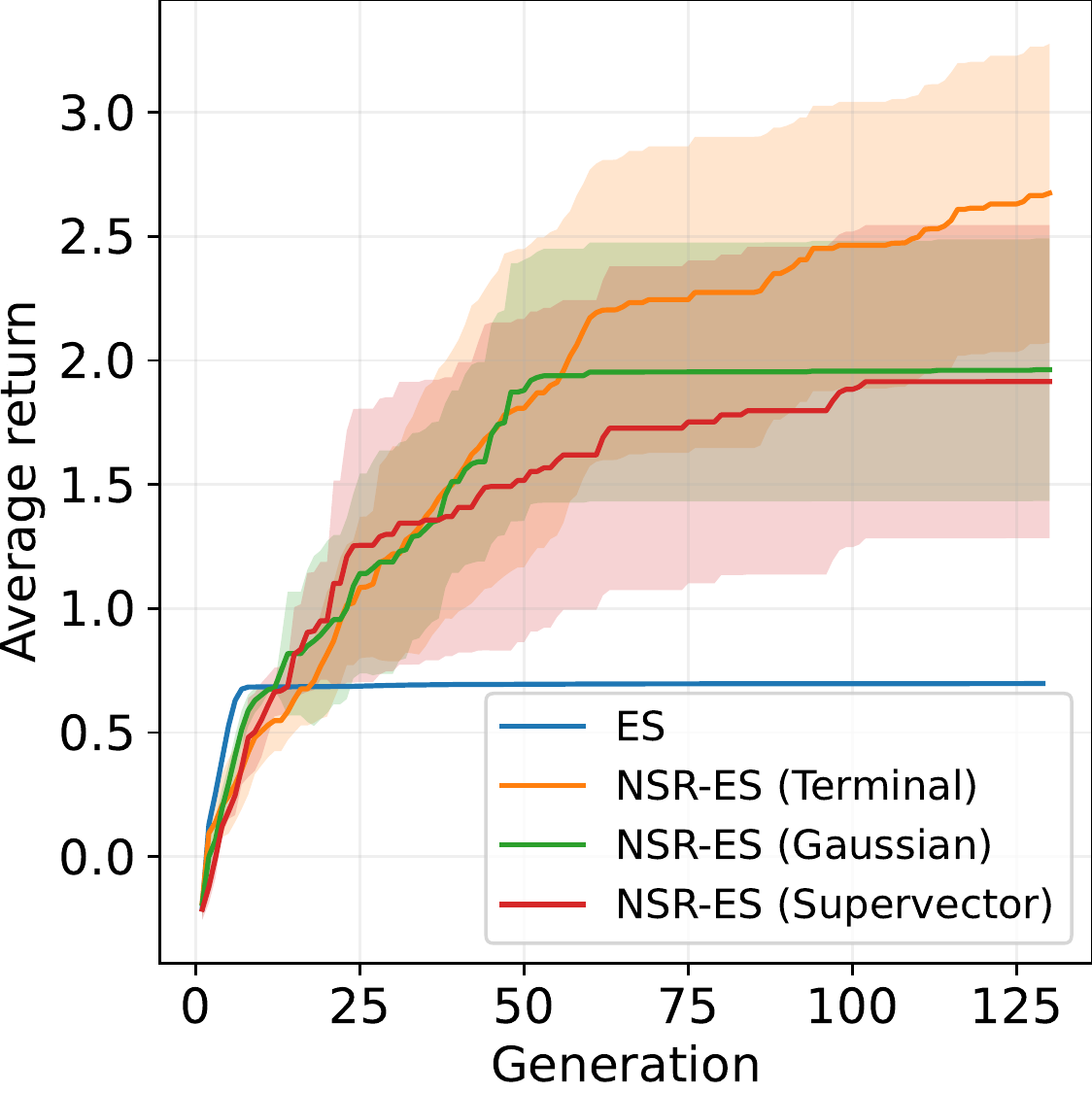}
                \caption{Novelty search}
                \label{fig:novelty-results}
            \end{subfigure}
            \caption{Results in novelty search (10 repetitions) and trust-region policy optimization (50 repetitions). Shaded region represents 95\% confidence interval.}
        \end{wrapfigure}

        Results in Figure \ref{fig:trust-region-results} indicate that action-based BC is best suited for this task, but both Gaussian and supervector approaches outperform no constraint. Gaussian BC slowly improves to the highest score overall, while supervector quickly reaches peak and then drops. We believe this is due to the non-gaussian nature of the state distribution, where supervector BC is able to move individual components on different modalities, allowing it to learn fast in the beginning but later become unstable. In the current form, state-based BCs are not practical as part of the training loop, but their advantage lies in the analysis of policies (Section~\ref{sec:evolution-of-policies}) and when encouraging novelty (Section~\ref{sec:novelty-search}).
        
    \subsection{Applying supervectors to novelty search}
        \label{sec:novelty-search}
        Novelty search encourages policies to find new behaviours to approach the task \cite{lehman2008exploiting}, which can take the form of maximizing BC distance between policies~\cite{conti2018improving}. While successful, previous work has used domain-specific heuristics like terminal states as BCs to describe the behaviour. Supervectors, or Gaussians, could be used instead as a more generalizable alternative. To test this, we construct a continuous 2D point environment similar to Pacchiano et al.~\cite{pacchiano2020learning}, where policy is rewarded for travelling to the positive y-direction. A wall prevents the policy from directly moving in this direction, and the policy has to learn to go around it. We use the methods and code of Conti et al.~\cite{conti2018improving}, where novelty encouragement is combined with fitness (NSR-ES) or agent is trained only for fitness (ES). Further details are in Appendix~\ref{sec:appendix-novelty}.
        
        Results in Figure \ref{fig:novelty-results} show that both single Gaussian and policy supervector help policy to overcome the obstacle. Using terminal state yields slightly better (but not significantly so) results, likely due to it being an \say{aligned} \cite{conti2018improving} BC with the objective, as walking away from previous locations can improve fitness. This demonstrates that supervectors (or even simple Gaussians) could be used as state-based BCs to encourage novel behaviour.
    
\section{Discussion and conclusions}
    As demonstrated, state-based BCs, and especially policy supervectors based on GMMs, can be used regardless of the environment while also scaling to thousands of policies. Compared to using returns or actions to define behaviour, state-based BC methods are more descriptive and capture dynamics of the environment into a behavioural description. These BCs can be used in the study of policies under different training algorithms, or to encourage exploration/novelty or impose limits on how much behaviour can change per policy update.
    
    While policy supervectors are applicable to studying policies as an outside observer, they are not yet practical as a part of the training loop: they require expensive sampling of the environment (as opposed to reusing already collected samples). This direction could be further explored by creating more sample efficient solutions or differentiable state-based BCs to train agents. Other topics include combining supervectors with the behavioural embeddings framework~\cite{pacchiano2020learning}, extending the novelty search experiments~\cite{parker2020effective} and further analysis of parameters learned by policy supervectors, such as the meaning of each adapted Gaussian component.

\begin{ack}
    We thank Ville Vestman for the help with UBM-GMM implementation, Jack Parker-Holder for insights on the behavioural embeddings and comments, and reviewers of the previous version of this work for providing constructive comments that helped us to improve this work. 
\end{ack}


\bibliography{references}
\bibliographystyle{abbrv}

\newpage

\appendix
\include{appendix/appendix}

\end{document}

%% file: appendix/appendix.tex

\section{Grid world illustration setup (Section \ref{sec:choosing-bc})}
    \label{sec:appendix-gridworld}
    Results in Figure \ref{fig:gridworld} of the original paper are the result of running the deterministic policies shown in the grid world images for ten thousand episodes. Agent receives a positive reward if it reaches the goal (higher the faster it reaches the goal with $+1$ being maximum), otherwise zero. Return-based distance is the absolute difference of average episodic rewards. Both action and state-based distances are computed as a sum of absolute differences between respective distributions. A policy is represented by tensor $\pi \in \mathbb R^{5 \times 5 \times 4}$, representing probability of choosing one of the four directions in each grid, and distance is between two such matrices is defined as $d_\text{action}(\pi, \hat \pi) = \sum_{i=1,j=1,k=1}^{5, 5, 4} |\pi_{i, j ,k} - \hat \pi_{i, j ,k}|$, summing over all grid and action pairs. Grids marked as walls are ignored from this. State distribution is estimated by running policy on the environment for ten thousand episodes and dividing visits to a cell by a total number of steps done during ten thousand episodes. The distance between two state distributions is then computed as $d_\text{state}(p_\pi, p_{\hat \pi}) = \sum_{i=1,j=1}^{5, 5} |p_{\pi}(i, j) - p_{\hat \pi}(i, j)|$. 

\section{Behavioural characterization evaluation experiment}
    \label{sec:appendix-evaluation}
    
    \subsection{Environments and training hyperparameters}
        Exact versions of the Gym environments are \verb|Acrobot-v0| (6D), \verb|Pendulum-v0| (3D), \verb|CartPole-v1| (4D), \verb|LunarLander-v2| (8D) and \verb|BipedalWalker-v3| (24D). The dimensionality refers to the observation vector dimensionality, which is also used as a state for the policy supervector extraction. We selected these environments to cover different levels of dimensionality and complexity while still remaining lightweight enough to allow us to scale experiments.
        
        For the PPO algorithm we use the implementation from stable-baselines \cite{stable-baselines} with hyperparameters provided by the rl-zoo package \cite{rl-zoo} (see Table \ref{tab:training-hyperparams}). In all environments, the agent reaches optimal reward with these settings. All agents use a neural network of two 64-unit layers with tanh-activations for value and policy estimation (separate network for both).

    \subsection{Evaluation metrics and data collection (Section \ref{sec:evaluation-metrics})}
        We train three PPO agents per environment and store $100$ checkpoints of the policy during training (a set of policies), and then collect $200$ trajectories of data per policy. We then sample $\{100, 50, 25, 10\}$ trajectories from these $200$, and repeat this sampling three times to simulate different runs of sampling of data. Three sets of policies per environment are used to capture stochasticity of final results over different sets of policies, while three repetitions of data sampling aim to capture stochasticity from sampling the environment. The axis over which we average for final results depends on the evaluation metric.
        
        \textbf{Correlation with return-distance.} For each policy we measure the average episodic reward over the $200$ trajectories and define return-distance as absolute difference between average returns $d_\text{returns} = |\mathbb E [R_{\pi_i}] - \mathbb E [R_{\pi_j}]|$, where $R_\pi$ is the sum of rewards of a trajectory. We then compute the Pearson correlation between this distance and the distance measured by the BC. The shown result is averaged over three sets of policies, three repetitions and over all $100 \times 100$ distance pairs, minus symmetric duplicates.
        
        \textbf{Distance error.} We compute distances between all policies inside a set. As policies remain fixed, and so does their behaviour, distances between them should stay the same relative to each other. If BC requires training, as is the case with our proposed methods, the scale of these numbers may change. For this reason, we normalize the distance matrices by min-max normalization to $[0, 1]$ over the whole distance matrix. We then select the first repetition with $100$ trajectories as \say{ground truth}, and for rest of the repetitions we measure the relative error to this distance $|d - d_\text{ground truth}| / d_\text{ground truth}$. The report result is averaged over the three sets of policies.
        
        \textbf{Distance variance.} We compute distance matrices as above, including normalization, and then compute the standard deviation over distance matrices over the three repetitions. The result shown is then an average of this over three sets of policies.
    
    \subsection{Implementation details of policy supervectors (Section \ref{sec:supervector})}
\begin{figure}[h]
        \centering
        \includegraphics[width=0.95\columnwidth]{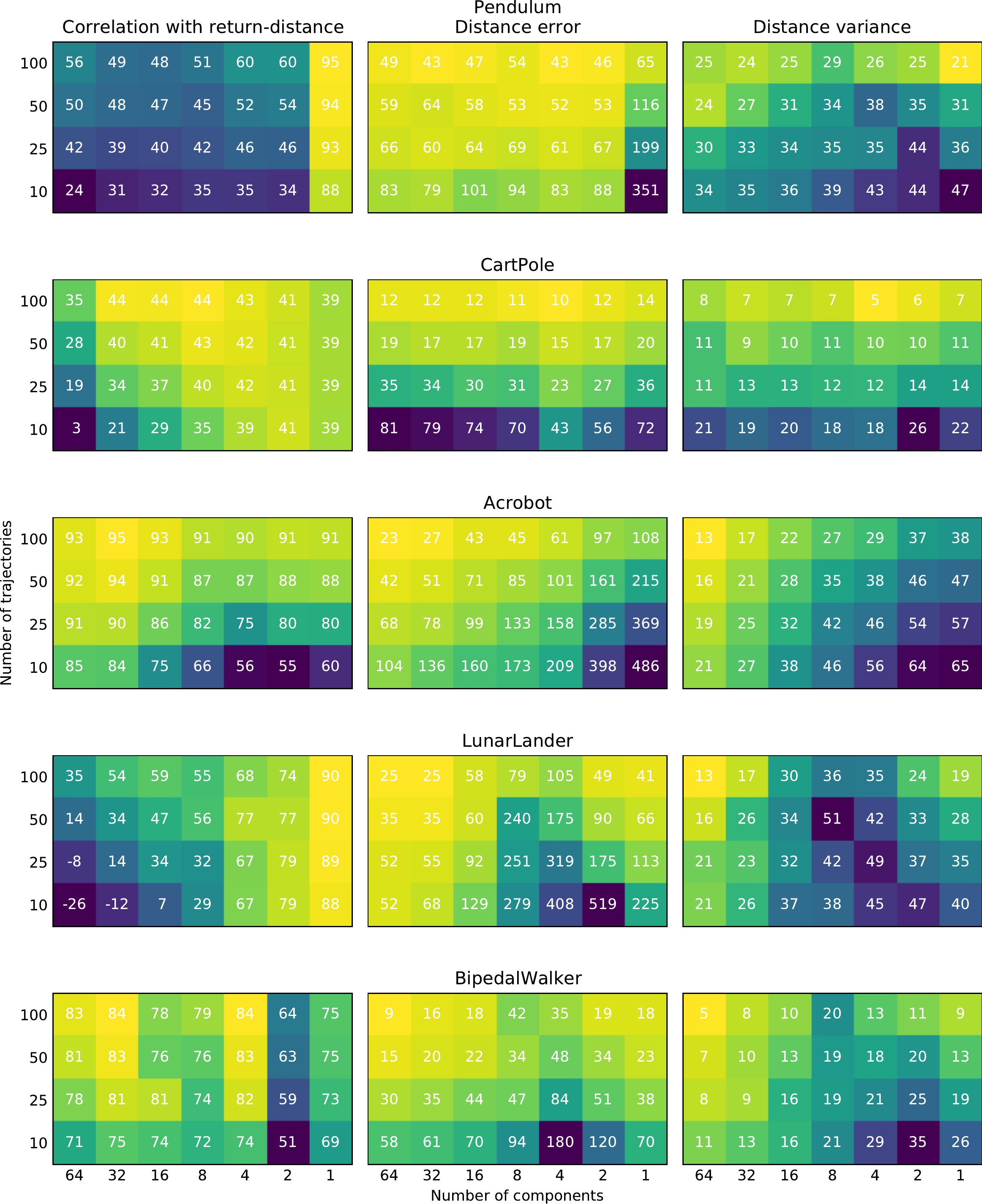}
        \caption{Brighter (yellow) colour means better. Per-environment results when analyzing the quality of final results under different amounts of data per policy and GMM components for policy supervectors. All values have been multiplied by a hundred.}
        \label{fig:ubm-comparison-appendix}
    \end{figure}

        See Algorithm \ref{algo:supervector} for pseudo-code of the method. We use scikit-learn to train GMMs \cite{scikit-learn}. We use relevance factor $r = 16$ in all of our experiments. GMMs use diagonal covariance matrices and are initialized with k-means before EM-training (the default setting). Figure~\ref{fig:ubm-comparison-appendix} repeats these experiments with different number of components, which was used to decide the $64$ components used in the main results (the lowest overall distance variance and error).
    
    \subsection{Implementation details of the discriminator method (Section \ref{sec:discriminator})}
        \label{sec:appendix-discriminator}
        See Algorithm \ref{algo:discriminator} for pseudo-code of the method. The network consists of three layers of $256$ units with tanh-activations, followed by a linear mapping to a scalar value. This value was clipped to interval $[-10, 10]$ (following Ni et al.~\cite{ni2020f}) and then fed through sigmoid function for $[0, 1]$ value. Given two datasets the discriminator is then trained to output high values for one and low values for the second one (see Section~4.2) for $30$ epochs with Adam optimizer (learning rate $10^{-3}$) with mini-batches of $128$ samples each.
    
    \subsection{Implementation details of the trajectory encoder method (Section \ref{sec:encoder})}
        \label{sec:appendix-encoder}
        See Algorithm \ref{algo:encoder} for pseudo-code of the method. Following Wang et al.~\cite{wang2017robust}, the encoder is a single bi-directional LSTM layer with $256$ units. For a single trajectory of states the backward output of this encoder is averaged over and then fed through a fully connected layer to parametrize a Gaussian to sample the embedding from (one fully connected layer for mean and another for variance). The decoder is a single fully connected layer that maps the sampled latent ($d = 256$) and previous state vector to a Gaussian (mean and variance), and finally the whole system is trained to maximize the log-likelihood of successive states in the dataset. In other words, the decoder needs to predict what will be the next state, given the previous state and a latent that (supposedly) describes the policy. After training a policy is represented by encoding all trajectories, sampling one latent each and finally fitting a diagonal, multivariate Gaussian on these latents, which is used to describe the policy.
        
        Training of the system is done with Adam optimizer \cite{adam} (learning rate $10^{-3}$) for five epochs over the data (data from all policies which are about to be compared), using mini-batches of eight trajectories at a time. We found five epochs to be sufficient to reach a stable loss, and eight trajectories per batch as one trajectory usually consists of hundreds of steps. Training these encoders took an order of magnitude longer than \textit{e.g.} discriminator method above, despite the number amount of epochs and small network size. On our four-core Intel Xeon system, the bipedal-walker experiments took four days alone.
    
    \subsection{Implementation details of the discretization method}
        For each policy we construct a matrix of $p_\pi = \mathbb R^{10 \times 10 \ldots D \ldots \times 10}$, where $D$ is the dimensionality of the state vector. For each state dimension we create $10$ uniformly spaced bins with edges at the observed minimum and maximum over all the data we are about to compare. We then discretize each state and increment corresponding cells in the $p_\pi$, and finally divide by the total number of states to obtain probability of being in that state over all states. Distance between two policies is measured as $d_\text{discretization}(p_{\pi_a}, p_{\pi_b}) = 0.5 \sum_s |p_{\pi_a}(s) - p_{\pi_b}(s)|$, where $s$ goes over all possible discretized states.

\section{Analysis of policies under different training algorithms (Section \ref{sec:evolution-of-policies})}
    \label{sec:appendix-algo-evaluation}
    \subsection{Experiment setup for t-SNE visualization}
        We trained five A2C agents, five PPO agents, one NEAT and one CMA-ES agent on the Gym environments used in the evaluation experiments. For A2C and PPO agents we use the same setup as above, and for NEAT/CMA-ES we use the code of Wann et al.~\cite{wann2019} with the hyperparameters shown in Table~\ref{tab:training-hyperparams}. We include only one run of NEAT/CMA-ES as they generate a large number of candidates. CMA-ES agents use fixed two-layer networks with sigmoid activations, with 40 units each for BipedalWalker and 5 units for other tasks. CMA-ES training terminates when fitness does not improve, which results in a smaller cloud of points for CMA-ES in t-SNE figures.
    
\begin{figure}[]
        \centering
        \includegraphics[width=0.95\columnwidth]{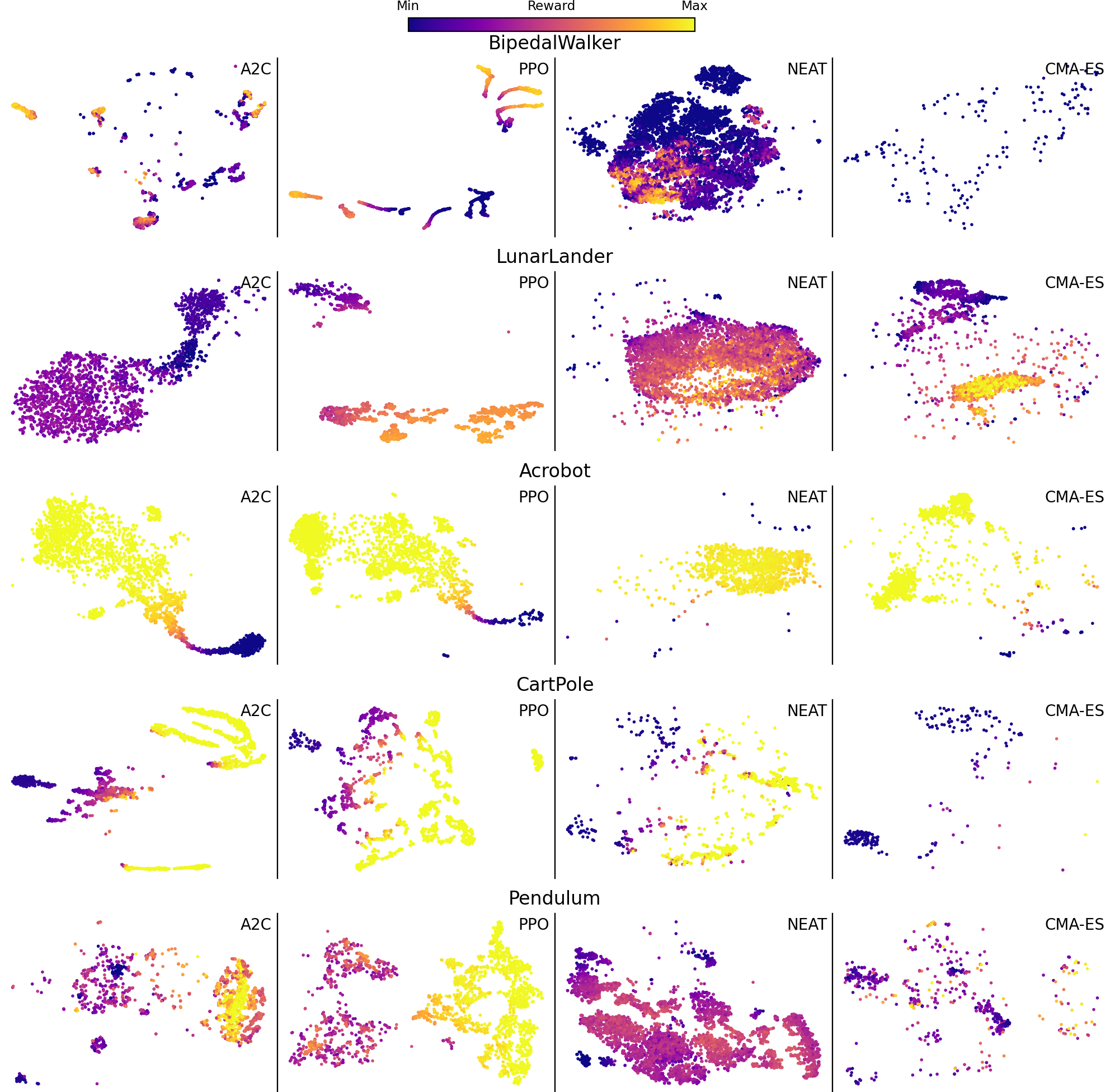}
        \caption{t-SNE plots of policies under different training algorithms in different environments. Each dot is a single policy (the behaviour characterization), where colour shows its average episodic return. A2C and PPO plots include policies from five different training runs, while CMA-ES and NEAT include policies from a single training run.}
        \label{fig:tsne}
    \end{figure}

        For RL agents we store $500$ checkpoints of policies per training run, and for NEAT/CMA-ES we store $25\%$ of the population candidates every tenth generation for visualization. We then train one UBM with $64$ components per environment, using at most $100$ policies per algorithm and/or $10$M samples in the dataset due to memory constraints. With the UBM we extract the policy supervectors and use the upper bound of KL-divergence as a distance for t-SNE, which is used to plot the results. See per-environment t-SNE visualizations in Figure~\ref{fig:tsne}.
        
    \subsection{Analysis of trust-region optimization}
        To study the effect of trust-region optimization on policy's behaviour during training, we train PPO agents with different ratio-clip values $\epsilon$, where the lower value should restrict updates to smaller changes in the policy. We train five PPO agents per different clip-ratio levels, where all other hyperparameters for training are the same (Table~\ref{tab:training-hyperparams}) except we set clip-ratio to the studied value. We also train five A2C agents with the original settings for comparison. We store $50$ policies during the training, gather $100$ trajectories per policy, train UBMs ($64$ components), extract policy-supervectors and compute the distance between successive policies experienced during training (a vector of $49$ values per one training run). We then sum these distances together for \say{total distance travelled}, average over the five repetitions and report the result. Results can be found in Table~\ref{tab:ppo-clip-results}, where we can see a connection between higher clip-ratio (larger trust-region) and the amount of distance travelled by the policy during training. This is especially pronounced in higher-dimensional environments, and also when comparing PPO (with trust-region) against A2C (no trust-region).
        
\begin{figure}[h]
        \centering
        \includegraphics[width=0.95\columnwidth]{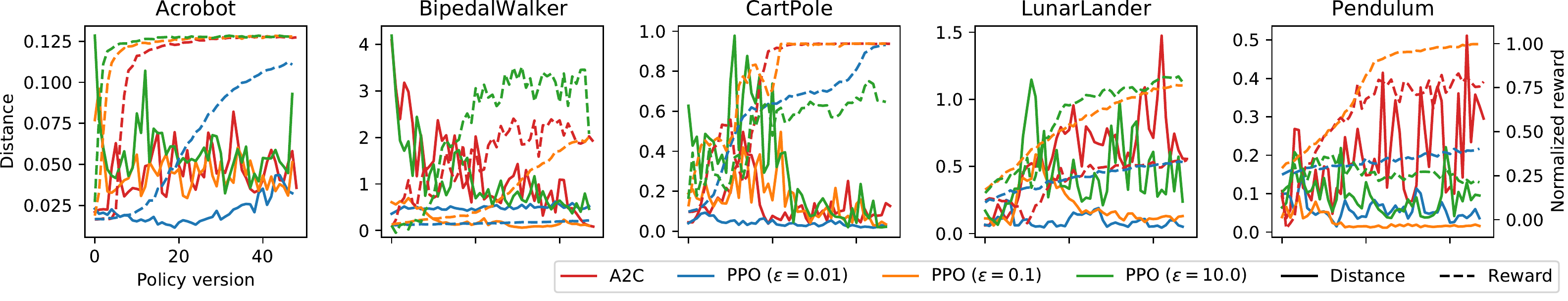}
        \caption{Learning curves with different clip-ratios $\epsilon$ when using PPO, with A2C included for comparison. Reward (dashed line) is normalized from minimum achievable reward (zero) to maximum (one). Distance refers to the distance between successive policy versions. Curves are averaged over five repetitions. Variance is omitted for visual clarity.}
        \label{fig:distances-appendix}
    \end{figure}

        To study how much change in behaviour is required as the agent gets better, we measure the Pearson correlation between average returns and distance to the next policy. Results in Figure~\ref{fig:distances-appendix} suggest that once policy gets better in the task, the amount of change to its behaviour decreases. Table~\ref{tab:ppo-clip-results} shows the final results, where we see a strong negative correlation between behaviour change and amount of returns, which concurs with the original observation. Results also show that the average distance travelled by policy increases as we increase the size of the trust-region, which concurs with the intuition of trust-region optimization.
    
    \subsection{Analysis of imitation learning}
        To study the connection between behaviour changes and imitation learning, we train five PPO expert agents on BipedalWalkerHardcore-v3 and LunarLander-v2 (hyperparameters in Table~\ref{tab:training-hyperparams}), collect $10$ trajectories of data and use it to train a behavioural cloning agent with $50$ epochs over the data. We chose a harder version of the bipedal-walker environment for a more challenging environment. We use PPO implementation from stable-baselines3 \cite{stable-baselines3} and imitation learning code from \say{imitation} library \cite{wang2020imitation}.
        
\begin{figure}[h]
        \centering
        \includegraphics[width=0.75\columnwidth]{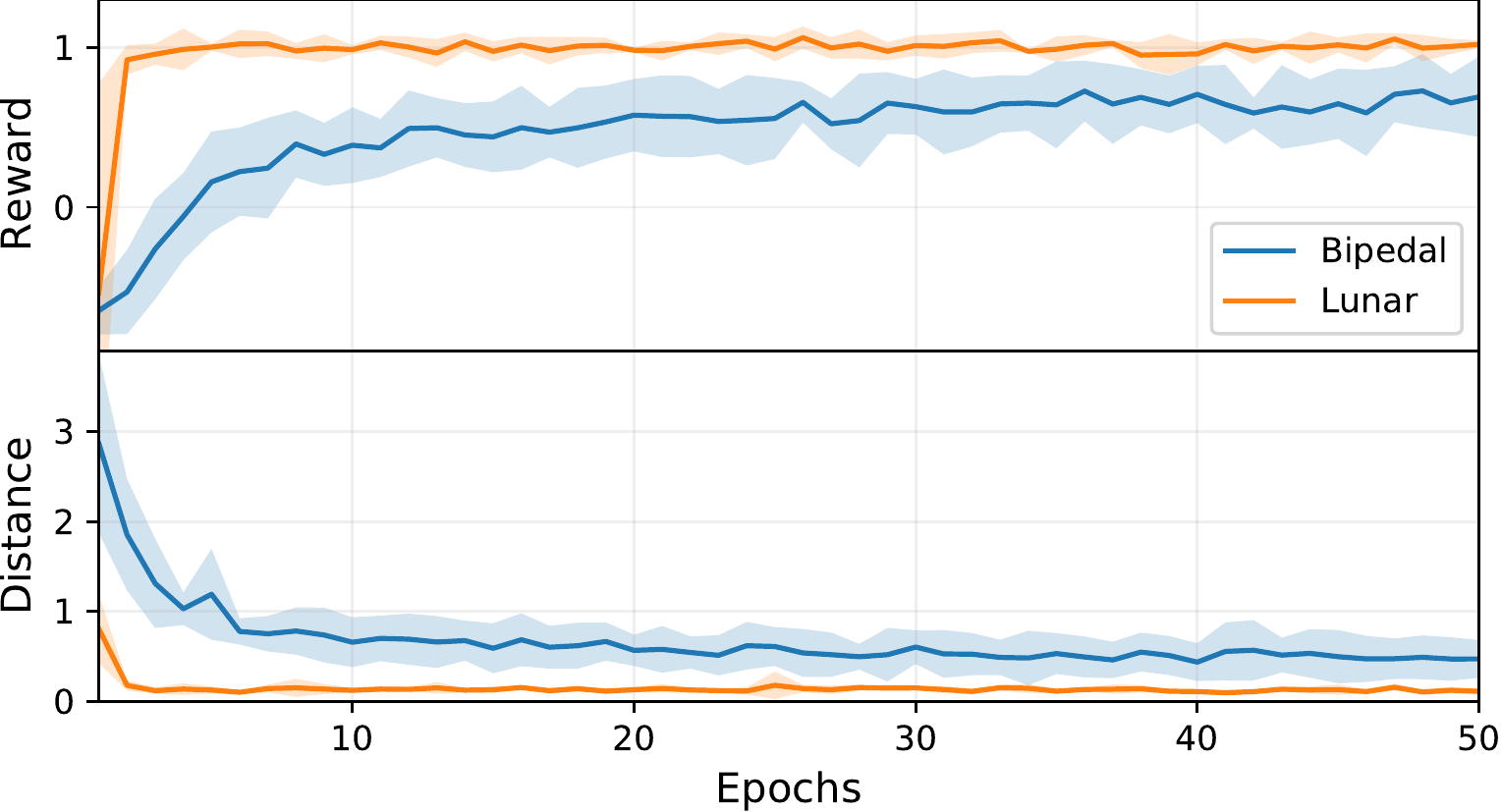}
        \caption{Behavioural cloning learning curves. The reward is normalized by dividing with expert's reward, with one equaling to performance of the expert. Distance refers to the distance to the expert policy. Averaged over five repetitions. The shaded area represents plus-minus one standard deviation.}
        \label{fig:bc-results}
    \end{figure}

        After each epoch, we store the current version of the trained policy. After training, we collect additional $100$ trajectories for the expert agent and for each BC policy, train the UBM ($64$ components), extract policy-supervectors and measure the distance of all BC policies to the expert agent. The resulting learning curve and distance-to-expert are shown in Figure~\ref{fig:bc-results}, where we see a connection between changes in returns and distance to expert policy, with an average Pearson correlation of $-0.465$ with a standard deviation of $0.297$. This shows that as behavioural cloning is training the policy to get a better average return (similar to expert's), its behaviour also approaches expert's behaviour.

\section{Novelty search experiment (Section \ref{sec:novelty-search})}
    \label{sec:appendix-novelty}
    The point-environment used in novelty search mimics the point environment used by Pacchiano et al.~\cite{pacchiano2020learning} but created in plain Python without MuJoCo. The agent lives on a 2D plane (no bounds), where observations are the current coordinates. Action space consists of a single real value $\phi \in [0, 2\pi]$ that represents the direction agent should move on to the next step (constant movement speed). Agent is rewarded based on movement on the y-axis directly with $r = y_\text{new} - y_\text{old}$. To move further in this axis, the agent has to avoid a trap by first moving either left or right on the x-axis (and slightly negative direction in the y-axis). If the agent does not avoid this trap it only gets $0.65$ reward, while the optimal the solution reaches $3.7$ reward. 
    
    For ES/NSR-ES implementations we used the code of Conti et al.~\cite{conti2018improving} and parameters from the humanoid experiments, but lowered data per update to $100$ episodes as the environment is significantly simpler than MuJoCo humanoid. We also decreased network size to two layers of $16$ units, followed by tanh-activations. For novelty search, we use a population of $K=3$ sets of parameters, with five episodes per BC (\textit{e.g.} Gaussian/supervector trained on five episodes of data, or terminal state is averaged over five episodes as was in the original implementation). For Gaussian BC we used symmetric KL-divergence (summed KL-divergence both ways), and for supervector, we used four Gaussian components and the KL-divergence upper bound distance.

\section{Applying supervectors for trust-region optimization (Section \ref{sec:trust-region})}
    \label{sec:appendix-trust}
    To test the usefulness of state-based trust-region in policy optimization, the environment (dubbed \say{dangerous path}) is a $N$-dimensional grid world, where the player always starts in the same location. The environment provides coordinates of the player as observations, and at each step, the agent may choose one of $N$ actions, which moves the player to the positive direction in that axis. For each location, there is only one valid action that moves the player forward and is rewarded with $+1$ reward. Two of the remaining actions are \say{mines} which will throw the player back to the beginning, and two actions do nothing. The assignment of these outcomes is done at random per grid. The agent is only rewarded when they move to a new grid location (\textit{i.e.} when they choose the correct action for the first time). In the experiments we set $N=5$ and limit episode length to $25$ steps, making the maximum reward $25$. We noticed the choice of the path can have a large effect on results, and for this reason, we ensure each compared method uses the same random seeds for the environment (\textit{i.e.} with four methods and ten repetitions each, we have ten random seeds used by all four methods).
    
    For the agent we modify the PPO implementation of stable-baselines3 \cite{stable-baselines3} by removing the ratio clipping and replacing this with a constraint check: if the constraint is exceeded, we stop updating the policy with current data and move on to collect trajectories for the next policy update. The constraint is checked for after every parameter update, comparing the old policy (which collected data) to the current one (after updates). We use eight environments and collect $512$ samples from each to perform policy updates to ensure a large amount of data to sample from. We do not use generalized advantage estimation and set the discount factor to zero as the agent is immediately rewarded for correct actions. The agent uses a two-layer network with $16$ units and tanh-activations each. The policy is updated for $100$ mini-batches of $64$ items over the collected data, or until constraint prevents updates.
    
    \textbf{No constraint.} Policy is always updated for $100$ mini-batches.
    
    \textbf{Total variation divergence (Max TV).} We measure total variation divergence of action probabilities over collected data and take the maximum of them. Essentially, this limits the maximum amount of how much probability of taking any single action (in any state) can change. Constraint value is searched from $\{0.001, 0.005, 0.01, 0.05, 0.1, 0.2, 0.3, 0.4, 0.5\}$.
    
    \textbf{Gaussian.} We collect five trajectories of data and fit a multivariate, diagonal Gaussian on it, and measure symmetric KL-divergence between old and current policy (summing both ways). We add a small amount of Gaussian noise ($\sigma=10^{-3}$) to collected states to ensure we have unique points. Constraint value is searched from $\{0.5, 1.0, 2.0, 3.0, 5.0, 10.0, 15.0, 20.0\}$.
    
    \textbf{Supervector.} As above, but we fit a four-component UBM on the data from old and new policy (five trajectories from both), extract policy supervectors and compute the upper bound of KL-divergence. We use four components due to the simplicity of the environment. Constraint value is searched from $\{0.01, 0.05, 0.1, 0.15, 0.2, 0.3, 0.4, 0.5\}$
    
    To ensure constraint violations happen we set the learning rate to relatively high $10^{-3}$ (for Adam optimizer, compared to default $3 \cdot 10^{-4}$). We then selected the constraint value with the largest area under the learning curve, which was averaged over ten repetitions, resulting in constraint values $0.4$, $10.0$ and $0.05$ for Max-TV, Gaussian and supervector, respectively. With these settings each of the methods terminate the policy updates occasionally, separating them from results with no constraint. With a lower learning rate we need a higher number of allowed updates than $100$, however, this started to become practically slow as Gaussian/supervector constraints constantly sampled the environment after every update. It is also worth noting that standard PPO with ratio-clipping $\epsilon = 0.2$ reached notably higher results than any of the compared results (above ten), which is due to more mini-batch updates per iteration ($640$ vs. $100$) and guided trust-region (the constraint is part of the loss).

\section{Hardware used to run the experiments}
    \label{sec:appendix-hardware}
    All experiments were run on either a four-core Intel Xeon W-2125, 64GB RAM workstation or on a 16-core Intel Xeon E5-2620 v4 (dual socket), 64GB RAM server machine. The only hard requirement for our experiments is the amount of system memory (64GB or more) and mass storage for storing policy trajectories (approx. 300GB in total), and with the above systems, all of the experiments can be conducted in approximately one week when parallelized over the two systems.

\begin{table}[t]
    \centering
    \caption{Min/Max values used for reward scaling per environment.}
    \label{tab:reward-scale}
    \begin{tabular}{ccc}
    \toprule
    \textbf{Environment} & \textbf{Min reward} & \textbf{Max reward} \\ \midrule
    Pendulum-v0       & -1600               & -200                \\
    Acrobot-v1        & -500                & -100                \\
    LunarLander-v2    & -230                & 200                 \\
    BipedalWalker-v3  & -100                & 300                 \\
    CartPole-v1       & 0                   & 500                 \\
    \bottomrule
    \end{tabular}
\end{table}

\begin{algorithm}[ht]
\caption{Gathering data for a policy (function \texttt{gatherData}). Python-like pseudo-code with OpenAI Gym environments.}
\label{algo:gather_data}
\begin{algorithmic}
    \STATE \textbf{Input:} An environment \textit{env}, a policy $\pi$ and number of trajectories to collect $M$.
    \STATE \textit{\#Initialize buffer.}
    \STATE $\bm B = [\;]$
    \FOR{$m \in \{1 \ldots M\}$}
        \STATE $s = \text{env.reset()}$
        \STATE $\text{done} = \text{False}$
        \WHILE {not done}
            \STATE $\bm B\text{.append}(\bm s)$
            \STATE $a \sim \pi(\bm s)$
            \STATE $\bm s, \text{done} = \text{env.step}(a)$
        \ENDWHILE
    \ENDFOR
    \STATE \textbf{Output:} Policy data $\bm B$.
\end{algorithmic}
\end{algorithm}

\begin{algorithm}[ht]
\caption{Computing distances with policy supervectors. Python-like pseudo-code.}
\label{algo:supervector}
\begin{algorithmic}
    \STATE \textbf{Input:} An environment \textit{env}, a set of policies we wish to study $\pi_n, n \in \{1 \ldots N\}$, number of GMM components $K$, number of trajectories to gather per policy $M$.
    \STATE \textit{\#Datasets of states for each policy.}
    \FOR{$n \in \{1 \ldots N\}$}
        \STATE $\bm B_n = \text{gatherData}(\text{env}, \pi_n, M)$
    \ENDFOR

    \STATE \textit{\#Concatenate all data into one big dataset.}
    \STATE $\bm B_\text{ubm} = \{\bm B_1, \ldots, \bm B_N\}$
    
    \STATE \textit{\#Fit a GMM with $K$ components (UBM)}.
    \STATE $(\bm \mu_\text{ubm}, \bm \Sigma_\text{ubm}, \bm w_\text{ubm}) = \text{gmm.fit}(\bm B_\text{ubm}, K)$
    
    \STATE \textit{\#Extract supervectors (adapted means)}
    \FOR{$n \in \{1 \ldots N\}$}
        \STATE \textit{\#MAP-adapt mean as in (1)-(5)}.
        \STATE $\bm{\hat \mu}_n = \text{mapAdapt}(\bm \mu_\text{ubm}, \bm \Sigma_\text{ubm}, \bm w_\text{ubm},\bm B_n)$
    \ENDFOR
    
    \STATE \textit{\#Initialize distance matrix}.
    \STATE $\text{policyDistances} = \bm 0^{N \times N}$
    \FOR{$i \in \{1 \ldots N\}$}
        \FOR{$j \in \{1 \ldots N\}$}
            \STATE \textit{\#Compute KL-divergence upper bound (6)}.
            \STATE $\text{policyDistances}_{i,j} = \text{KLUpperBound}(\bm{\hat \mu}_i, \bm{\hat \mu}_j, \bm \mu_\text{ubm}, \bm \Sigma_\text{ubm}, \bm w_\text{ubm})$
        \ENDFOR
    \ENDFOR
        
    \STATE \textbf{Output:} Distance matrix $\text{policyDistances}$
\end{algorithmic}
\end{algorithm}

\begin{algorithm}[ht]
\caption{Computing distances with discriminators. Python-like pseudo-code.}
\label{algo:discriminator}
\begin{algorithmic}
    \STATE \textbf{Input:} An environment \textit{env}, a set of policies we wish to study $\pi_n, n \in \{1 \ldots N\}$, number of GMM components $K$ and number of trajectories to gather per policy $M$.

    \STATE \textit{\#Datasets of states for each policy.}
    \FOR{$n \in \{1 \ldots N\}$}
        \STATE $\bm B_n = \text{gatherData}(\text{env}, \pi_n, M)$
    \ENDFOR

    \STATE \textit{\#Initialize distance matrix}.
    \STATE $\text{policyDistances} = \bm 0^{N \times N}$
    \FOR{$i \in \{1 \ldots N\}$}
        \STATE \textit{\#Train discriminator to separate $i$\textsuperscript{th} dataset from all others}.
        \STATE $D_i = \text{trainDiscriminator}(\bm B_i, \bm B_{j \neq i})$
        \FOR{$j \in \{1 \ldots N\}$}
            \STATE \textit{\#Test how likely is that $j$\textsuperscript{th} dataset came from $i$\textsuperscript{th} policy using (10)}.
            \STATE $\text{distance} = \frac{1}{|\bm B_j|} \sum_{\bm s \in \bm B_j} f_i(\bm s) $
            \STATE \textit{\#Accumulate both ways for symmetry}.
            \STATE $\text{policyDistances}_{i,j} = \text{policyDistances}_{i,j} + \text{distance}$
            \STATE $\text{policyDistances}_{j,i} = \text{policyDistances}_{j,i} + \text{distance}$
        \ENDFOR
    \ENDFOR
    \STATE \textbf{Output:} Distance matrix $\text{policyDistances}$
\end{algorithmic}
\end{algorithm}

\begin{algorithm}[ht]
\caption{Computing distances with trajectory encoders. Python-like pseudo-code.}
\label{algo:encoder}
\begin{algorithmic}
    \STATE \textbf{Input:} An environment \textit{env}, a set of policies we wish to study $\pi_n, n \in \{1 \ldots N\}$, number of GMM components $K$ and number of trajectories to gather per policy $M$.
    \STATE \textit{\#Collect trajectories for each policy (keep trajectories separate).}
    \FOR{$n \in \{1 \ldots N\}$}
        \STATE $\bm T_n = [\;]$
        \FOR{$m \in \{1 \ldots M\}$}
            \STATE $\text{states} = [\;]$    
            \STATE $s = \text{env.reset()}$
            \STATE $\text{done} = \text{False}$
            \WHILE {not done}
                \STATE $\text{states.append}(\bm s)$
                \STATE $a \sim \pi(\bm s)$
                \STATE $\bm s, \text{done} = \text{env.step}(a)$
            \ENDWHILE
            \STATE $\bm T_{n}\text{.append(states)}$
        \ENDFOR
    \ENDFOR

    \STATE \textit{\#Train a variational autoencoder to reconstruct trajectories, using all data}.
    \STATE $\text{encoder} = \text{trainVAE}(\bm T)$
    
    \STATE \textit{\#Encode and sample latents from each trajectory and fit one diagonal Gaussian per policy}.
    \FOR{$n \in \{1 \ldots N\}$}
        \STATE $\bm Z = [\;]$
        \FOR{$t \in \{1 \ldots |\bm T_n|\}$}
            \STATE $\bm z \sim \text{encoder}(\bm T_{n, t})$
            \STATE $\bm Z\text{.append}(\bm z)$
        \ENDFOR
        \STATE $\bm \mu_n = \bm Z\text{.mean(axis=0)}; \; \bm \sigma_n = \bm Z\text{.std(axis=0)}$
    \ENDFOR

    \STATE \textit{\#Initialize distance matrix}.
    \STATE $\text{policyDistances} = \bm 0^{N \times N}$
    \FOR{$i \in \{1 \ldots N\}$}
        \FOR{$j \in \{1 \ldots N\}$}
            \STATE \textit{\#Compute KL-divergence between policys' Gaussians}.
            \STATE $\text{distance} = \text{DiagonalGaussianKL}(\bm \mu_i, \bm \mu_j, \bm \sigma_i, \bm \sigma_j)$
            \STATE \textit{\#Accumulate both ways for symmetry}.
            \STATE $\text{policyDistances}_{i,j} = \text{policyDistances}_{i,j} + \text{distance}$
            \STATE $\text{policyDistances}_{j,i} = \text{policyDistances}_{j,i} + \text{distance}$
        \ENDFOR
    \ENDFOR
    \STATE \textbf{Output:} Distance matrix $\text{policyDistances}$
\end{algorithmic}
\end{algorithm}

     \begin{table}[]
            \caption{Agent training hyperparameters. ``*" marks parameters that are linearly decayed to zero over training. "Bipedal.H.`` refers to BipedalWalkerHardcore, used in the imitation learning experiments with PPO expert agent.}
            \label{tab:training-hyperparams}
            \centering
            \small
            \begin{tabular}{lllllll}
\toprule
\textbf{PPO parameters}         & \textbf{Bipedal.H.} & \textbf{Bipedal.} & \textbf{LunarLan.} & \textbf{Acrobot} & \textbf{CartPole} & \textbf{Pendulum} \\ \midrule
Environment steps       & 100 000 000              & 5 000 000              & 1 000 000            & 1 000 000        & 100 000           & 2 000 000         \\
Number of envs.          & 16                       & 16                     & 16                   & 16               & 8                 & 8                 \\
Rollout size                    & 2048                     & 2048                   & 1024                 & 256              & 32                & 2048              \\
Training epochs                 & 10                       & 10                     & 4                    & 4                & 20                & 10                \\
Batch size                      & 1024                     & 1024                   & 512                  & 512              & 256               & 512               \\
Entropy weight             & 0.001                    & 0.001                  & 0.01                 & 0.0              & 0.0               & 0.0               \\
Policy ratio clip               & 0.2*                & 0.2                    & 0.2                  & 0.2              & 0.2*         & 0.2               \\
Value ratio clip                & 0.2*                & 0.2                    & 0.2                  & 0.2              & 0.2*         & 0.2               \\
Learning rate              & 0.00025*            & 0.00025                & 0.00025              & 0.00025          & 0.001*       & 0.0003            \\
GAE $\lambda$                   & 0.95                     & 0.95                   & 0.98                 & 0.94             & 0.8               & 0.95              \\
Discount factor                 & 0.99                     & 0.99                   & 0.999                & 0.99             & 0.98              & 0.99              \\ \\
\textbf{A2C parameters}         &                          &                        &                      &                  &                   &                   \\ \midrule
Environment steps       & -                        & 5 000 000              & 200 000              & 500 000          & 500 000           & 2 000 000         \\
Number of envs          & -                        & 16                     & 8                    & 16               & 8                 & 8                 \\
N-steps                         & -                        & 5                      & 5                    & 5                & 5                 & 5                 \\
Entropy weight             & -                        & 0.0                    & 0.00001              & 0.0              & 0.0               & 0.0               \\
Learning rate           & -                        & 0.0007*           & 0.00083*        & 0.0007           & 0.0007            & 0.0007            \\
Discount factor                 & -                        & 0.99                   & 0.995                & 0.99             & 0.99              & 0.95              \\ \\
\textbf{NEAT parameters} &                          &                        &                      &                  &                   &                   \\ \midrule
Number of gens.           & -                        & 1024                   & 1024                 & 256              & 256               & 256               \\
Population size                 & -                        & 192                    & 128                  & 128              & 128               & 128 
            \\ \\
\textbf{CMA-ES parameters} &                          &                        &                      &                  &                   &                   \\ \midrule
Number of gens.           & -                        & 1024                   & 1024                 & 256              & 256               & 256               \\
Population size                 & -                        & 192                    & 128                  & 128              & 128               & 128 \\
\bottomrule     
\end{tabular}
        \end{table}
        
    
    \begin{table}[]
        \caption{Total distance travelled by the policy during PPO training with different ratio-clipping values $\epsilon$, averaged over five training runs along with standard deviation. $\text{Corr}(R, d)$ is the Pearson correlation between average episodic reward and distance to next policy in training.}
        \label{tab:ppo-clip-results}
        \centering
        \begin{tabular}{cccccc}
        \toprule
$\epsilon$            & \textbf{Bipedal} & \textbf{Lunar} & \textbf{Acro} & \textbf{Cart} & \textbf{Pendulum} \\ \midrule 
& \multicolumn{5}{c}{Distance traveled} \\ \midrule
.01                   & 23.6$\pm$1.1                      & 4.5$\pm$.5                      & 1.0$\pm$.1                     & 2.0$\pm$.1                     & 2.8$\pm$.4                      \\
.1                    & 1.7$\pm$1.5                       & 9.9$\pm$.5                      & 2.2$\pm$.1                     & 7.7$\pm$1.6                    & 1.2$\pm$.1                      \\
.2                    & 11.3$\pm$2.0                      & 9.7$\pm$1.1                     & 2.6$\pm$.2                     & 13.6$\pm$1.0                   & 1.3$\pm$.2                      \\
.3                    & 12.3$\pm$4.5                      & 1.8$\pm$.4                      & 2.5$\pm$.3                     & 12.2$\pm$2.0                   & 1.6$\pm$.3                      \\
.4                    & 11.9$\pm$2.3                      & 14.7$\pm$2.3                    & 2.4$\pm$.3                     & 14.9$\pm$1.2                   & 2.1$\pm$.5                      \\
.5                    & 12.0$\pm$2.6                      & 15.0$\pm$.7                     & 2.6$\pm$.1                     & 15.3$\pm$1.7                   & 2.5$\pm$.4                      \\
.75                   & 17.7$\pm$7.4                      & 16.7$\pm$1.0                    & 2.5$\pm$.3                     & 15.9$\pm$3.7                   & 3.7$\pm$.5                      \\
1                     & 25.1$\pm$1.9                      & 18.9$\pm$2.5                    & 2.6$\pm$.2                     & 14.0$\pm$4.0                   & 4.1$\pm$.6                      \\
2                     & 19.0$\pm$3.4                      & 17.3$\pm$1.2                    & 2.7$\pm$.2                     & 17.1$\pm$4.6                   & 3.5$\pm$.8                      \\
5                     & 25.4$\pm$7.2                      & 22.9$\pm$4.3                    & 2.9$\pm$.3                     & 2.2$\pm$2.6                    & 3.4$\pm$.3                      \\
10                    & 51.5$\pm$21.1                     & 21.5$\pm$2.6                    & 2.8$\pm$.2                     & 14.9$\pm$4.9                   & 4.5$\pm$.7                      \\
A2C                   & 61.8$\pm$16.6                     & 29.4$\pm$4.3                    & 2.2$\pm$.2                     & 6.5$\pm$.7                     & 9.8$\pm$2.4                     \\ \\
 & \multicolumn{5}{c}{$Corr(R, d)$}                                                                                                                                       \\ \midrule
.01                   & .25$\pm$.18                       & -.06$\pm$.12                    & .46$\pm$.11                    & -.54$\pm$.07                   & -.13$\pm$.20                    \\
.1                    & -.47$\pm$.19                      & -.25$\pm$.03                    & -.34$\pm$.23                   & -.14$\pm$.11                   & -.53$\pm$.04                    \\
.2                    & -.73$\pm$.08                      & -.34$\pm$.05                    & -.53$\pm$.21                   & -.01$\pm$.10                   & -.56$\pm$.11                    \\
.3                    & -.61$\pm$.07                      & -.45$\pm$.11                    & -.43$\pm$.14                   & -.20$\pm$.14                   & -.33$\pm$.10                    \\
.4                    & -.66$\pm$.07                      & -.50$\pm$.05                    & -.40$\pm$.10                   & -.10$\pm$.09                   & -.42$\pm$.11                    \\
.5                    & -.58$\pm$.12                      & -.43$\pm$.10                    & -.35$\pm$.13                   & .06$\pm$.24                    & -.19$\pm$.17                    \\
.75                   & -.54$\pm$.11                      & -.32$\pm$.09                    & -.40$\pm$.23                   & -.14$\pm$.28                   & .08$\pm$.09                     \\
1                     & -.60$\pm$.20                      & -.27$\pm$.06                    & -.37$\pm$.12                   & .23$\pm$.30                    & .11$\pm$.10                     \\
2                     & -.67$\pm$.10                      & -.15$\pm$.10                    & -.45$\pm$.22                   & -.15$\pm$.23                   & .04$\pm$.28                     \\
5                     & -.63$\pm$.14                      & .03$\pm$.13                     & -.47$\pm$.19                   & -.28$\pm$.17                   & .19$\pm$.18                     \\
10                    & -.58$\pm$.13                      & .03$\pm$.14                     & -.37$\pm$.18                   & .04$\pm$.39                    & .20$\pm$.26                     \\
A2C                   & -.46$\pm$.25                      & .40$\pm$.10                     & .20$\pm$.08                    & -.30$\pm$.12                   & .20$\pm$.18 \\ \bottomrule
\end{tabular}
    \end{table}
    
